\documentclass[journal]{IEEEtran}
\usepackage{amssymb}
\usepackage{color}
\usepackage{multirow}
\usepackage{mathrsfs}
\usepackage{flushend}

\newtheorem{definition}{Definition}

\usepackage{cite}
\ifCLASSINFOpdf
\usepackage[pdftex]{graphicx}
\graphicspath{{../pdf/}{../jpeg/}}
\DeclareGraphicsExtensions{.pdf,.jpeg,.png}
\else
\usepackage[dvips]{graphicx}
\graphicspath{{../eps/}}
\DeclareGraphicsExtensions{.eps}
\fi
\usepackage[cmex10]{amsmath}
\usepackage{array}
\usepackage[tight,footnotesize]{subfigure}
\usepackage{stfloats}

\usepackage{arydshln}
\usepackage{rotating}
\usepackage{microtype}
\IEEEoverridecommandlockouts
\usepackage{epstopdf}

\begin{document}
\title{An MDL-Based Classifier for Transactional~Datasets with Application in Malware Detection}
\author{\IEEEauthorblockN{Behzad Asadi and Vijay Varadharajan}\\
\IEEEauthorblockA{The University of Newcastle, Australia}\vspace{-20pt}}

\maketitle

\begin{abstract}
We design a classifier for transactional datasets with application in malware detection. We build the classifier based on the minimum description length (MDL) principle. This involves selecting a model that best compresses the training dataset for each class considering the MDL criterion. To select a model for a dataset, we first use clustering followed by closed frequent pattern mining to extract a subset of closed frequent patterns (CFPs). We show that this method acts as a pattern summarization method to avoid pattern explosion; this is done by giving priority to longer CFPs, and without requiring to extract all CFPs. We then use the MDL criterion to further summarize extracted patterns, and construct a code table of patterns. This code table is considered as the selected model for the compression of the dataset. We evaluate our classifier for the problem of static malware detection in portable executable (PE) files. We consider API calls of PE files as their distinguishing features. The presence-absence of API calls forms a transactional dataset. Using our proposed method, we construct two code tables, one for the benign training dataset, and one for the malware training dataset. Our dataset consists of 19696 benign, and 19696 malware samples, each a binary sequence of size 22761. We compare our classifier with deep neural networks providing us with the state-of-the-art performance. The comparison shows that our classifier performs very close to deep neural networks. We also discuss that our classifier is an interpretable classifier. This provides the motivation to use this type of classifiers where some degree of explanation is required as to why a sample is classified under one class rather than the other class.

\end{abstract}
\IEEEpeerreviewmaketitle
\vspace{-0pt}
\section{Introduction}
Machine learning algorithms are being widely deployed in different applications as automated decision-making tools due to their generalization capabilities. This includes malware detection which is our main concern in this work. Traditional algorithms for malware detection search for known signatures which requires them to have a copy of all malware samples. These algorithms are not effective nowadays as (i) polymorphism is used within a malware family, (ii) the number of new malware families is increasingly growing, and (iii) they are not capable of zero-day malware detection. Machine learning algorithms are good candidates for automated malware detection. This is because they can extract complex patterns using different attributes of a malware, and also they can help with zero-day malware detection as they can generalize to new samples~\cite{MLforMalware}.

Malware detection can be divided into two main categories of dynamic (behavioral) and static (code) malware detection. In dynamic malware detection, samples are executed, and their run-time behavior is monitored to create indicators of malicious activities. In this type of malware detection, malware samples can easily adapt their run-time behavior to evade detection when they are aware of the normal behavior. In static malware detection, binary codes of samples are examined without executing them to create indicators of malicious activities. In this work, we consider malware detection in portable executable (PE) files using static analysis. Different types of features have been used for static malware detection in PE files such as API calls~\cite{FGSMMalware}, byte-level N-grams~\cite{StaticNgram}, features from the PE header~\cite{StaticHeader}, and a combination of different types of features~\cite{EMBER}. We consider API calls of PE files to distinguish between malware and benign samples. The presence-absence of API calls forms a transactional dataset. Transactional datasets can also appear in other applications such as healthcare where a sample represents a set of symptoms of a patient, marketing where a sample represents a basket of items purchased by a customer, and natural language processing where a sample represents a set of terms in a document.

A major challenge in designing classifiers for malware detection is that malware authors actively try to evade anti-malware systems by modifying existing malware samples without changing their functionalities. Therefore, the final aim is to develop a classifier that not only has a high performance considering available samples, but is also robust to functionality-preserving modifications. The first step towards having a robust classifier is to understand why a sample is classified under one class rather than the other one. This helps to develop a better understanding of how the model works, which in turn can be used to make the model more robust to evasion attacks. Therefore, interpretability is an important property for a classifier to improve its robustness. An interpretable model should be able to provide some good explanations to its users as to why a specific decision is made regarding a given sample. Several properties are considered for a good explanation to a user~\cite{InterpretableMLBook}. One important property is that explanations need to be contrastive. This means that a user is interested to know not only why a decision is made but also why not another decision. Another important property is that explanations need to be succinct. This means that they need to provide a short list of important reasons for a decision rather a complete list of reasons. There are also several other important motivations for having an interpretable machine learning model. One is that interpretable models help us to understand the scenarios where these models fail. This is important where wrong decisions by a model have serious consequences. For instance, classifying a malware sample as benign can bring the whole critical infrastructure down. Another important motivation is to detect biases in models. This can happen when a model is trained on a biased training dataset. Having an interpretable model helps us to extract these biases by looking at the reasons for its decisions.

In this work, we design an MDL-based classifier as a type of intrinsically interpretable models. Intrinsic in the sense that the model itself is interpretable due to its structure. The design involves selecting a model that best describes (compresses) the training dataset for each class considering the MDL criterion. The MDL principle has already been used for both classification and anomaly detection in transactional datasets\cite{KRIMP, OddOneOut,SLIM,KRIMPCategorical}. However, our proposed model-selection method is able to handle much larger datasets (more than 10 to 100 times~larger).

We search among code tables of patterns as the family of models for the compression of transactional datasets. Code table construction can also be viewed as a pattern summarization problem aimed at selecting a small interesting subset of a large list of candidate patterns. The MDL criterion selects a subset of patterns that best describes the dataset under consideration, i.e., shortest possible description. The code table of selected patterns is considered as the selected model by the MDL criterion~\cite{KRIMP}. In transactional datasets, closed frequent pattern mining (CFPM)~\cite{FPMBook} is used to created the large list of candidate patterns required for code table construction. Frequent patterns are sets of items occurring together more than a user-decided threshold. Closed frequent patterns (CFPs) are those frequent patterns that do not have a superset with the same number of occurrence. In large datasets, having a large threshold causes obvious and short patterns, and having a small threshold can cause pattern explosion which makes both pattern mining, and code table construction computationally very expensive. 

We therefore propose an MDL-based model-selection method for large transactional datasets. In our method, we first employ clustering to divide our dataset into a number of clusters. We then select a subset of clusters based on a criterion proposed to determine the quality of a cluster. We next perform CFPM in only high-quality clusters separately. The outputs of CFPM for all high-quality clusters are merged as the final output of CFPM. This approach extracts a subset of all CFPs for our dataset. We show that our approach helps to avoid pattern explosion by considering priority for longer CFPs, and without requiring to extract all CFPs. We finally use the MDL criterion to further summarize extracted patterns, and construct a code table of patterns as the selected model.

We utilize our classifier for static malware detection in PE files using a dataset consisting of 19696 benign, and 19696 malware samples each a binary sequence of size 22761 (representing 22761 unique API calls in our dataset). We compare our classifier with deep neural networks providing us with the state-of-the-art performance. The comparison shows that our classifier performs very close to neural networks. We also discuss about the interpretability of our classifier, and how it can help to understand why a sample is classified under one class rather than the other class.

\textit{Organization:} This work is organized as follows. In Section~\ref{Sec:TransactionalDataset}, we present some preliminaries for transactional datasets. In Section~\ref{Sec:mdl}, we describe the MDL principle and its applications for pattern summarization, classification and anomaly detection. In Section~\ref{Sec:ProposedMethod}, we present our proposed method for MDL-based model selection. In Section~\ref{Sec:Performance}, we show the advantages of our proposed method for model selection, and also compare our classifier with deep neural networks. In Section~\ref{Sec:Interpretability}, we discuss about the interpretability of our classifier. In Section~\ref{Sec:Conclusion}, we conclude the work. In Appendix, we review two algorithms that we use for CFPM, and clustering.

\section{Transactional Datasets}\label{Sec:TransactionalDataset} 
In this section, we present some preliminaries for transactional datasets, and also outline some issues related to CFPM and clustering for these datasets.

\subsection{Preliminaries} 
Assuming that a dataset consists of $m$ possible items, $\mathcal{I}=\left\{1,2,3,\ldots,m\right\}$ represents the set of all items. The~whole dataset, denoted by $\mathcal{D}$, is a non-empty multiset (bag) of transactions, i.e., $\mathcal{D}=\{\mathcal{T}_1,\mathcal{T}_2,\ldots,\mathcal{T}_n\}$, where each transaction is a subset of $\mathcal{I}$, i.e., $\mathcal{T}_j\subseteq \mathcal{I}$.  We say that a transaction $\mathcal{T}_j$ supports an itemset $\mathcal{P}$ (which is also a subset of $\mathcal{I}$) if $\mathcal{P}\subseteq \mathcal{T}_j$. The support of an itemset, denoted by $sup(\cdot)$, is the number of transactions that support the itemset. Considering $\mathcal{D}(\mathcal{P})$ as the multiset of transactions that support the itemset $\mathcal{P}$, and $\mathcal{D}(\mathcal{Q})$ as the multiset of transactions that support the itemset $\mathcal{Q}$, we therefore have
\begin{itemize}
	\item $sup(\mathcal{P})=\left|\mathcal{D}(\mathcal{P})\right|$,
	\item $\mathcal{D}(\mathcal{P}\cup \mathcal{Q})=\mathcal{D}(\mathcal{P})\cap \mathcal{D}(\mathcal{Q})$,
	\item If $\mathcal{P}\subseteq\mathcal{Q}$, then $\mathcal{D}(\mathcal{P})\supseteq\mathcal{D}(\mathcal{Q})$,
	\item If $\mathcal{P}\subseteq\mathcal{Q}$, then $sup(\mathcal{P})\geq sup(\mathcal{Q})$,
\end{itemize}
where $|\cdot|$ denotes the cardinality of a multiset. 

\subsection{Closed Frequent Pattern Mining}\label{Sec:CFPM}

We here address the problem of CFPM in transactional datasets\cite{FPMBook}. An itemset is frequent if its support is greater than or equal to a user-decided threshold, denoted by $minsup$. A frequent itemset is closed if it has no superset with the same support. We employ the Linear Time Closed Itemset Mining (LCM) algorithm~\cite{LCM} to directly extracts CFPs. This is as opposed to first extracting all frequent patterns (via algorithms such as the Apriori algorithm~\cite{Apriori}), and then selecting the subset of CFPs. The LCM algorithm can dramatically reduce the computation time when the number of frequent patterns is exponentially larger than the number of CFPs. Refer to Appendix for an overview of the LCM algorithm

\subsection{Clustering}\label{Sec:Clustering}

We here address the problem of clustering for transactional datasets. As traditional clustering algorithms using a pair-wise similarity do not perform well for transactional datasets, several algorithms have been developed for these datasets which consider a global criterion function~\cite{LargeItemClustering,clope}. Using a global criterion function has also this advantage that the user does not need to know the number of clusters in advance. The global criterion function is defined such that intra-cluster similarity is maximised, and inter-cluster similarity is minimised. 

In this work, we employ the Clustering with sLOPE (CLOPE) algorithm~\cite{clope} which is a fast and scalable algorithm for clustering transactional datasets. In this algorithm, we do not need to know the number of clusters in advance. The two parameters of this algorithm are repulsion factor, used to control intra-cluster similarity, and maximum cluster number, used to provide an upper limit for the number of clusters. Refer to Appendix for an overview of the CLOPE algorithm

\section {MDL Principle and Its Applications}\label{Sec:mdl}
In this section, we present the MDL principle, and its applications for pattern summarization, classification, and anomaly detection.
\subsection {MDL Principle}
Kolmogorov complexity theory, also known as algorithmic information theory, was developed to measure the information in objects in isolation, i.e., without knowing the distribution underlying the object. As in data mining, we normally do not know the underlying distribution of our data, we use algorithmic information theory to measure the information in our data. The Kolmogorov complexity of an object is the descriptive complexity of the object, which is the length of the shortest computer program that can describe the object. This is formally defined as follows~\cite{ITBook}. 

\begin{definition}
The Kolmogorov complexity of an object $x$ with respect to a universal computer $\mathcal{U}$, denoted by  $K_{\mathcal{U}}(x)$, is defined as
\begin{align*}
K_{\mathcal{U}}(x)=\underset{prog:\mathcal{U}(prog)=x}{\min} \ell(prog),
\end{align*}
which is the minimum length over all programs that print $x$ and halt.
\end{definition}

However, we cannot compute the Kolmogorov complexity of an object. Therefore, in practice, the MDL principle is utilized. Using the crude MDL criterion, we choose a model from a family of models, $\mathcal{M}$, that minimizes the two-term objective function $\ell(x\mid M_i)+\ell(M_i)$ where $\ell(x\mid M_i)$ is the number of bits required to describe the object given the model, and $\ell(M_i)$ is the number of bits required to describe the model itself. Hence, using the crude MDL criterion, we have
\begin{align*}
\ell_\text{best}(x)=\underset{M_i\in\mathcal{M}}{\min}\; (\ell(x\mid M_i)+\ell(M_i)).
\end{align*}


\subsection{MDL-based Pattern summarization}\label{Sec:CodeTable}
The MDL principle can be used for pattern summarization where we want to select a small subset of an existing large set of candidate patterns denoted by $\mathcal{F}$. In this part, we present the algorithm proposed by Vreeken et al.~\cite{KRIMP} which uses the MDL principle for pattern summarization. This algorithm performs pattern summarization by searching among code tables of patterns as the family of models to describe the data. A code table, denoted by $CT$, has two columns: the first column consists of selected patterns, and the second column consists of binary codes used to encode the patterns in the first column. This algorithm, which basically outputs a semi-adaptive  compression dictionary, selects the best code table as
\begin{align}\label{mdlct}
CT_{\text{best}}=\underset{CT}{\arg\min}\; (\ell(\mathcal{D}\mid CT)+\ell(CT)).
\end{align}

In the algorithm, as the search space for constructing code tables is very large, a heuristic approach is used to select the best code table. This heuristic approach consists of three steps. In the first step, candidate patterns in the set $\mathcal{F}$ are ordered descending first by their support, second by their length. In the second step, a standard code table consisting of all singleton items is constructed. In the third step, candidate patterns from the ordered $\mathcal{F}$ are examined one by one. In this step, if adding a candidate pattern to the current code table results in a smaller objective function, i.e., $\ell(\mathcal{D}\mid CT)+\ell(CT)$, it is kept in the code table, otherwise it is dropped. This leads to keeping only a small subset of $\mathcal{F}$ in the final code table. The patterns in the final code table are considered as the patterns chosen by the MDL principle.

We here explain how the two terms $\ell(\mathcal{D}\hskip-2pt\mid\hskip-2pt CT)$ and $\ell(CT)$ in equation~\eqref{mdlct} are calculated. The first term in equation~\eqref{mdlct}, $\ell(\mathcal{D}\mid CT)$, is calculated as
\begin{align*}
\ell(\mathcal{D}\mid CT)=\sum_{\mathcal{T}\in\mathcal{D}}\ell(\mathcal{T}\mid CT)=\sum_{\mathcal{T}\in\mathcal{D}}\sum_{\mathcal{P}\in \mathcal{C}(\mathcal{T})}\ell(\mathcal{P}\mid CT),
\end{align*}
where $\ell(\mathcal{P}\mid CT)$ is the length of the binary code for the pattern $\mathcal{P}$, and $\mathcal{C}(\mathcal{T})$ is the set of patterns used to cover $\mathcal{T}$. The patterns covering a transaction satisfy the following properties
\begin{align*}
\forall\;\mathcal{P}_i,\mathcal{P}_j\in \mathcal{C}(\mathcal{T}),\text{ if } \mathcal{P}_i\neq\mathcal{P}_j \text{ then } \mathcal{P}_i\cap\mathcal{P}_j=\emptyset,
\end{align*}
and 
\begin{align*}
\bigcup_{\mathcal{P}\in \mathcal{C}(\mathcal{T})}\mathcal{P}=\mathcal{T}.
\end{align*}
As there can be several ways (different sets of patterns) to cover a transaction, the patterns in the code table are ordered descending first by their length, next by their support; the patterns are selected according to this order to cover a transaction.  

The lengths of binary codes in the second column of the code table, i.e., $\ell(\mathcal{P}\mid CT)$, are determined by the Shannon code which is a prefix code. The more a pattern used in the cover of transactions, the shorter its code. Therefore, by defining the usage of a pattern $\mathcal{P}$ as
\begin{align*}
\mathit{usage}(\mathcal{P})=\left|\left\{\mathcal{T}\in\mathcal{D}:\mathcal{P}\subseteq \mathcal{C}(\mathcal{T})\right\}\right|, 
\end{align*}
the code for the pattern $\mathcal{P}$ is of length
\begin{align*}
\ell(\mathcal{P}\mid CT)&=\left \lceil  -\log Pr(\mathcal{P}\mid D)\right \rceil\\
&=\left \lceil-\log \left(\frac{usage(\mathcal{P})}{\sum_{\mathcal{P'}\in CT}usage(\mathcal{P'})}\right)\right \rceil.
\end{align*}

The second term in equation~\eqref{mdlct}, $\ell(CT)$, is calculated as
\begin{align}\label{lengthcodetabel}
\ell(CT)&= \sum_{i\in\mathcal{I}} n_i\log (|\mathcal{I}|+1)+\sum_{\mathcal{P}\in CT}\log (|\mathcal{I}|+1)\nonumber\\&\hskip12pt+\sum_{\mathcal{P}\in CT}\ell(\mathcal{P}\mid CT),
\end{align}
where $n_i$ is the number of times that item $i$ appears in the patterns in the first column of the code table. The number of all possible items in first column of the code table considering a separator between each two patterns is $|\mathcal{I}|+1$. The first two terms on the left-hand side of equation~\eqref{lengthcodetabel} correspond to encoding the first column of the code table. The last term on the left-hand side of equation~\eqref{lengthcodetabel} corresponds to encoding the second column of the code table consisting of prefix binary codes.

\subsubsection{Example}
We here provide an example for pattern summarization. In this example, we consider the following dataset which consists of five items and 10 transactions. 
\begin{center}
	\begin{tabular}{ccccc}
		1&  2&  3& 4 & 5\\
		\hline
		1&  1&  1& 1& 0\\
		1&  1&  1& 1& 0\\ 
		1&  1&  0& 1& 0\\  
		0&  1&  1& 1& 1\\ 
		0&  0&  1& 1& 1\\
		0&  0&  0& 1& 1\\ 
		0&  1&  0& 0& 0\\
		0&  0&  1& 0& 0\\
		0&  0&  0& 1& 0\\
		0&  0&  0& 0& 1
	\end{tabular}
\end{center}
Each row represents a transaction. This dataset can be represented as
\begin{align*}
\mathcal{D}&=\left\{\{1,2,3,4\}^2,\{1,2,4\},\{2,3,4,5\},\right.\\
&\hskip90pt\left.\{3,4,5\},\{4,5\},\{2\},\{3\},\{4\},\{5\}\right\},
\end{align*}
where $\mathcal{D}$ is a multiset, and the superscript for an element shows the multiplicity of that element. We use CFPM with $minsup=1$ to extract all CFPs of this dataset. This is to form the list of candidate patterns required to construct an MDL-based code table for this dataset. Using extracted CFPs, the ordered list of candidate patterns is

\begin{center}
	\begin{tabular}{c|c}
		$\mathcal{P}$&  $sup(\mathcal{P})$  \\
		\hline
		$\{4\}$&  7  \\ 
		$\{3\}$&  5  \\  	
		$\{2\}$&  5\\
		$\{3,4\}$&  4  \\  	
		$\{2,4\}$&  4\\
		$\{5\}$&  4  \\ 
		$\{2,3,4\}$&  3  \\  	
		$\{1,2,4\}$&  3\\
		$\{4,5\}$&  3  \\ 
		$\{1,2,3,4\}$&  2  \\  	
		$\{3,4,5\}$&  2  \\ 
		$\{2,3,4,5\}$&  1
	\end{tabular}
\end{center}

The final code table using the described approach is

\begin{center}
	\begin{tabular}{|c|c|}
		\hline 
		$\mathcal{P}$ & binary code length\\  
		\hline 
		$\left\{1,2,4\right\}$   &  $3$  \\ 
		\hline 
		$\left\{4\right\}$     &  $3$  \\ 
		\hline 
		$\left\{3\right\}$     &  $2$  \\
		\hline 
		$\left\{2\right\}$     &  $4$  \\	
		\hline
		$\left\{5\right\}$     &  $3$  \\	
		\hline
		$\left\{1\right\}$     &  $8$  \\	
		\hline
	\end{tabular} 
\end{center}
which shows the effectiveness of the MDL principle for pattern summarization. In the second column of the code table, we have provided the lengths of binary codes than binary codes themselves. This is because the lengths are important than the codes themselves. Note that item 1 does not appear in the cover of any transactions, i.e., its usage is equal to zero. We keep all singleton items in the final code table by giving them a small usage when their usage is zero. This is to be able to cover any unseen transactions. 

\subsection{MDL-based Classifier}
We here explain how to utilize the MDL principle to build a binary classifier. Supervised learning consists of two phases of training and test. In the training phase, we select a model that best describes the training dataset of each class using the MDL criterion
\begin{align*}
M_{\mathcal{D}_1}&=\underset{M_i\in\mathcal{M}}{\arg\min}\; (\ell(\mathcal{D}_1\mid M_i)+\ell(M_i)),\\
M_{\mathcal{D}_2}&=\underset{M_i\in\mathcal{M}}{\arg\min}\; (\ell(\mathcal{D}_2\mid M_i)+\ell(M_i)).
\end{align*}

In the test phase, if for a transaction $\mathcal{T}$, we have   
\begin{align*}
\ell(\mathcal{T}\mid M_{\mathcal{D}_2})<\ell(\mathcal{T}\mid M_{\mathcal{D}_1}),
\end{align*}
this implies that
\begin{align*}
Pr(\mathcal{T}\mid \mathcal{D}_2)> Pr(\mathcal{T}\mid \mathcal{D}_1).
\end{align*}
Consequently, we classify the sample $\mathcal{T}$ under the second class. Otherwise, we classify it under the first class. Note that the term $\ell(M_i)$ in the crude MDL criterion prevents the model to be overfitted during the training phase.

\subsection{MDL-based Anomaly Detector}
We here explain how to utilize the MDL principle to build an anomaly detector.
In anomaly detection, we assume that we have access to only a dataset of normal samples (possibly with some small numbers of anomalies which have been mislabelled as normal samples). Therefore, we just select a model that best describes the normal dataset, $\mathcal{D}$, using the MDL criterion
\begin{align*}
M_{\mathcal{D}}=\underset{M_i\in\mathcal{M}}{\arg\min}\; (\ell(\mathcal{D}\mid M_i)+\ell(M_i)).
\end{align*}

Hence, if for the two sample $\mathcal{T}_1$ and $\mathcal{T}_2$, we have  
\begin{align*}
\ell(\mathcal{T}_1\mid M_{\mathcal{D}})<\ell(\mathcal{T}_2\mid M_{\mathcal{D}}),
\end{align*}
this implies
\begin{align*}
P(\mathcal{T}_1\mid \mathcal{D})>P(\mathcal{T}_2\mid \mathcal{D}).
\end{align*}

This says that the larger $\ell(\mathcal{T}\mid M_{\mathcal{D}})$, the smaller $P(\mathcal{T}\mid \mathcal{D})$. Therefore, in this method, we need to define a threshold $\theta$ using which we say that a sample is anomaly~if 
\begin{align*}
\ell(\mathcal{T}\mid M_\mathcal{D})> \theta.
\end{align*}

\begin{figure}[t]
	\centering
	\includegraphics[width=0.35\textwidth]{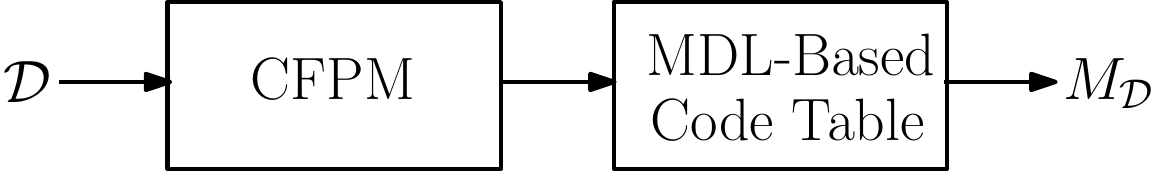}
	\caption{The two-step KRIMP method for MDL-based model selection.} 
	\label{Fig:TwoStepMethod}
	\vspace{-10pt}
\end{figure}

\section{Proposed Method for Model Selection}\label{Sec:ProposedMethod} 
In this work, we aim to construct a classifier for large transactional datasets based on the MDL principle. To do so, as mentioned earlier, we need to select a model that best describes the training dataset for each class using the MDL criterion. This can be done using the two-step KRIMP algorithm proposed by Vreeken et al.~\cite{KRIMP}, shown in Fig.~\ref{Fig:TwoStepMethod}. In this algorithm, CFPM is first used to extract all CFPs. These CFPs are then considered as candidates patterns to construct a code table of patterns as described in Section~\ref{Sec:CodeTable}. The code table is considered as the selected model by the MDL criterion. However, for large datasets, this algorithm can be computationally very expensive. In the first step, having a small $minsup$ in CFPM can lead to pattern explosion, and consequently extracting all CFPs is computationally very expensive. In the second step, we need to test each extracted pattern to decide whether to keep the pattern in the final code table or to drop the pattern. This step is also computationally expensive, and very slow. This is because this step needs to be done for all extracted CFPs in a specific serial order (it cannot be parallelized), and also the whole dataset is used for testing each pattern. To address these problems, we propose using clustering in conjunction with CFPM. We show that this approach extracts a subset of all CFPs by giving priority to longer CFPs. Here, we first explain how clustering affects CFPM. We then present our MDL-based model-selection method.

\subsection{CFPM after Clustering}

We consider applying a CFPM algorithm to the clusters of a dataset separately, and then taking the union over the outputs of the CFPM algorithm for clusters. This provides us with a subset of all CFPs for the whole dataset (i.e., the output of directly applying a CFPM algorithm to the whole dataset). This is because if a pattern is a CFP considering one of the clusters, it is also a CFP considering the whole dataset. We here discuss that this method can be considered as a pattern summarization method that gives priority to longer patters. This is important for our application as the compression is mainly achieved through longer patterns.  

To cluster the whole dataset, we utilize a clustering algorithm designed for transactional datasets as described in Section~\ref{Sec:Clustering}. The clustering algorithm tries to group similar transactions into one cluster, and dissimilar ones into separate clusters. This implies that the longer a pattern supported by several transactions, the higher the probability that clustering groups those transactions into one cluster. After clustering, four types of clusters can exist corresponding to a CFP: Type-I cluster where the support of the pattern is zero; Type-II cluster where the support of the pattern is non-zero but less than $minsup$; Type-III cluster where the support of the pattern is greater than $minsup$, but there is a superset for the pattern with the same support; and Type-IV cluster where the support of the pattern is greater than $minsup$, and there is no superset for the pattern with the same support. Therefore, if we do not have any Type-IV clusters corresponding to a CFP, clustering leads to dropping that pattern. We divide cases leading to dropping a CFP into two scenarios. In the first scenario, all clusters are Type-I or Type III. In the second scenario, there is at least one Type-II cluster. In both scenarios, we do not have any Type-IV clusters.

\begin{figure*}[t]
	\centering
	\includegraphics[width=0.82\textwidth]{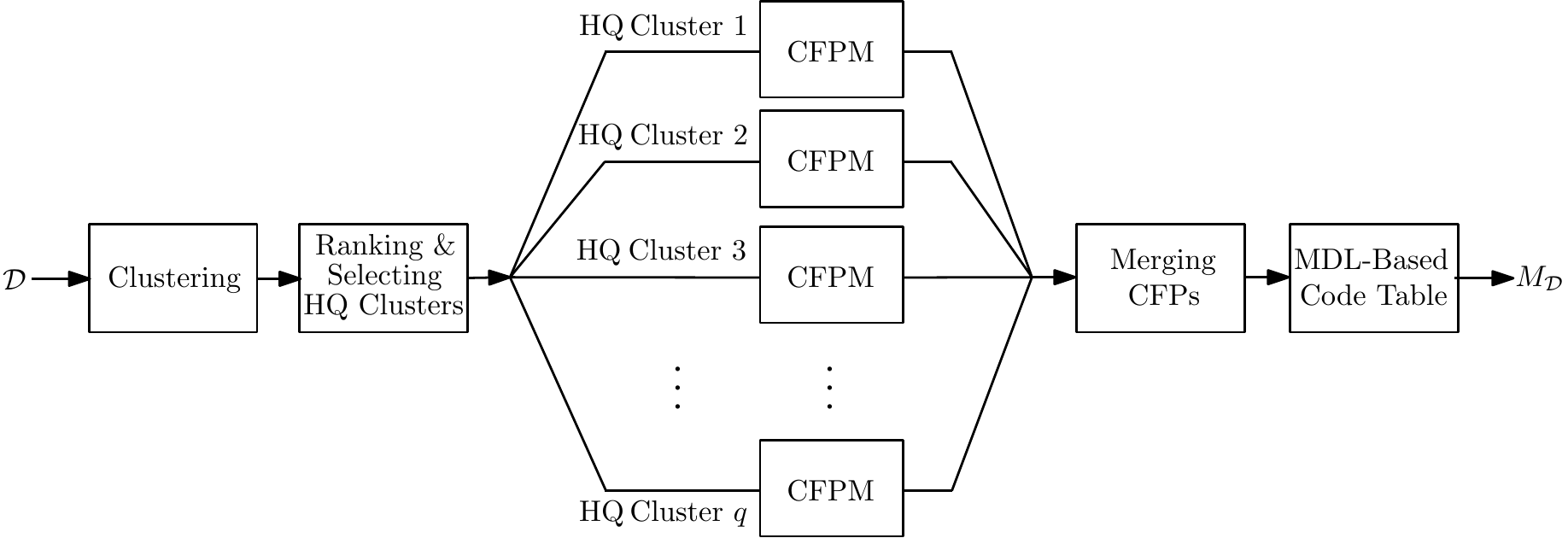}
	\caption{Proposed method for MDL-based model selection.} 
	\label{Fig:ProposedMethod}
\end{figure*}

We here show these two scenarios via two examples. In these two examples, we use the CLOPE algorithm with repulsion factor equal to four, and maximum cluster number equal to two. We also consider $minsup$ to be two. 

In this first example where we face the first scenario, our dataset consists of five items and seven transactions as shown here
\begin{center}
	\begin{tabular}{ccccc}
		1&  2&  3& 4 & 5\\
		\hline
		1&  1&  1& 1& 0\\ 
		1&  1&  1& 1& 0\\  
		0&  1&  1& 1& 1\\ 
		0&  1&  1& 1& 1\\
		0&  0&  1& 1& 1\\ 
		0&  0&  1& 1& 1\\
		0&  0&  0& 1& 1
	\end{tabular}
\end{center}
which is represented as
\begin{equation*}
\mathcal{D}=\left\{\{1,2,3,4\}^2,\{2,3,4,5\}^2,\{3,4,5\}^2,\{4,5\}\right\}.
\end{equation*}
The CFPs of this dataset are
\begin{align*}
CFP_\text{w}&=\left\{\{1,2,3,4\},\{2,3,4\},\{2,3,4,5\},\right.\\
&\hskip90pt\left.\{3,4,5\},\{3,4\},\{4\},\{4,5\}\right\}.
\end{align*}
After clustering, the following two clusters are formed
\begin{align*}
\mathbb{C}_1&=\left\{\{1,2,3,4\}^2\right\},\\
\mathbb{C}_2&=\left\{\{2,3,4,5\}^2,\{3,4,5\}^2,\{4,5\}\right\},
\end{align*}
and the union of CFPs for these two clusters~is
\begin{equation*}
CFP_\text{c}=\left\{\{1,2,3,4\},\{2,3,4,5\},\{3,4,5\},\{4,5\}\right\}.
\end{equation*}
For any of the missing CFPs, both clusters are Type-III.

In this second example where we face the second scenario, our dataset consists of nine items and four transactions
\begin{center}
	\begin{tabular}{ccccccccc}
				1&  2&  3& 4& 5& 6 & 7 & 8& 9\\
				\hline
				0&  0&0&  0&  0& 0& 1& 1 & 1\\ 				  
				0&  0&0&  1&  1& 1& 1& 1 & 1\\ 
				1&  1&1&  1&  1& 1& 0& 0 & 0\\  
				1&  1&1&  0&  0& 0& 0& 0 & 0\\
	\end{tabular}
\end{center}
which is represented as
\begin{equation*}
\mathcal{D}=\left\{\{7,8,9\},\{4,5,6,7,8,9\},\{1,2,3,4,5,6\},\{1,2,3\}\right\}.
\end{equation*}
The CFPs of this dataset are
\begin{align*}
CFP_\text{w}&=\left\{\{7,8,9\},\{4,5,6\},\{1,2,3\}\right\}.
\end{align*}
After clustering, these two clusters are formed
\begin{align*}
\mathbb{C}_1&=\left\{\{7,8,9\}\right\},\\
\mathbb{C}_2&=\left\{\{4,5,6,7,8,9\},\{1,2,3,4,5,6\},\{1,2,3\}\right\},
\end{align*}
and the union of CFPs for these two clusters~is
\begin{align*}
CFP_\text{c}&=\left\{\{4,5,6\},\{1,2,3\}\right\}.
\end{align*}
It can be seen the pattern $\{7,8,9\}$ is dropped as the result of clustering. Both clusters are Type-II for this pattern. 

The patterns dropped as the result of facing the first scenario are those which are formed from the intersection of longer CFPs. In both scenarios, the patterns dropped as the result of clustering are mainly from shorter patterns. That is why we consider CFPM after clustering as a pattern summarization method which gives priority to longer patterns. 

\subsection{Proposed Model-Selection Method}

In this section, we explain our proposed MDL-based model-selection method shown in Fig.~\ref{Fig:ProposedMethod}. 

As discussed in the last section, we use clustering in conjunction with CFPM to extract a subset of CFPs by giving priority to longer patterns. The maximum number of clusters can be decided based on the parameter $minsup$. The larger the parameter $minsup$, the smaller the number of clusters. For a large $minsup$, we do not face pattern explosion, and consequently we do not need clustering. We use this method when we have a small $minsup$, and as a result we face pattern explosion. This is to directly avoid pattern explosion, i.e., not by first extracting all CFPs, and then selecting a subset of them. As our target is to minimize the probability that a long CFP is dropped, and also maximize pattern summarization, we propose the following strategy. We first cluster the dataset, and rank clusters according to the following criterion
\begin{equation}\label{clusterquality}
Quality(\mathbb{C}_i)=\frac{H(\mathbb{C}_i)}{\left|\mathbb{C}_i\right|},
\end{equation}
where $H(\mathbb{C}_i)$ and $\left|\mathbb{C}_i\right|$ are the height and the number of transactions of cluster~$i$ respectively. The height of cluster $\mathbb{C}_i$ is defined as
\begin{equation*}
H(\mathbb{C}_i)=\frac{\sum_{\mathcal{T}_j\in \mathbb{C}_i}|\mathcal{T}_j|}{|\cup_{\mathcal{T}_j\in \mathbb{C}_i}\mathcal{T}_j|}.
\end{equation*}
The cluster quality takes a value between zero and one. It is equal to one where all the transactions of a cluster are the same (the highest quality). We next select a subgroup of clusters as high-quality (HQ) clusters by setting a quality threshold, and perform CFPM in only HQ clusters. In HQ clusters, transactions share majority of their items, and as a result the number of CFPs in these clusters is not large even by considering a small $minsup$. Low-quality (LQ) clusters are the main reason for pattern explosion, and the output of CFPM in these clusters consists of mainly short patterns.

As the output of the pattern-mining stage, we take the union over the outputs of CFPM in HQ clusters. We finally construct a code table of patterns according to Section~\ref{Sec:CodeTable} as the selected model.

\section{Performance Evaluation}\label{Sec:Performance}
In this section, we first compare our proposed model-selection method with the two-step KRIMP method using the small mushroom dataset. This is to show the advantages of our proposed model-selection method in pattern summarization, and constructing a code table. We then evaluate our classifier on a large dataset of API calls for static malware detection in PE files. We also compare our classifier with deep neural networks providing us with the state-of-the-art performance in static malware detection. 

\subsection{Small Dataset}
We use the mushroom dataset to compare our model-selection method with the KRIMP method. The mushroom dataset, which is a categorical dataset, consists of 4208 edible samples, and 3916 poisonous samples. After converting the dataset into a transactional dataset using one-hot encoding, each sample of the dataset is a binary sequence of size 117.

\subsubsection{KRIMP Method} We here use the KRIMP method, shown in Fig.~\ref{Fig:TwoStepMethod}, to construct a code table, $CT_\text{e}$, for the edible dataset, $\mathcal{D}_\text{e}$, and a code table, $CT_\text{p}$, for the poisonous dataset, $\mathcal{D}_\text{p}$. We first use the LCM algorithm for CFPM. Considering $minsup$ to be 0.5 percent of the dataset size for each class (i.e., $minsup=21.04$ for the edible dataset, and $minsup=19.6$ for the poisonous dataset), we have 34781 CFPs in the edible dataset, and 24041 CFPs in the poisonous dataset. Using extracted CFPs as candidate patterns, we then construct two code tables for the edible, and the poisonous datasets. The final code table for the edible dataset consists of 238 patterns, and the one for the poisonous dataset consists of 176 patterns. Using constructed code tables via the KRIMP method, we have
\begin{align*}
\ell(\mathcal{D}_\text{e}\mid CT_\text{e})+\ell(CT_\text{e})=182190\;\; bits,\\
\ell(\mathcal{D}_\text{p}\mid CT_\text{p})+\ell(CT_\text{p})=177215\;\; bits,
\end{align*}
which show the compression achieved for the edible, and the poisonous datasets. Before the compression, the edible dataset consists of $4208\times117=492336$ bits, and the poisonous dataset consists of $3916\times117=458172$ bits.
\subsubsection{Proposed Method} We here use our proposed method, shown in Fig.~\ref{Fig:ProposedMethod}, to construct a code table for the edible dataset, and a code table for the poisonous dataset. To cluster the dataset for each class, we use the CLOPE algorithm with repulsion factor equal to four, and maximum cluster number equal to eight. This provides us with eight clusters for the edible dataset, and six clusters for the poisonous dataset. The cluster qualities of the edible dataset are 0.73, 0.71, 0.71, 0.67, 0.28, 0.63, 0.73, and 0.76. We consider all edible clusters as HQ clusters. The cluster qualities of the poisonous dataset are 0.71, 0.71, 0.65, 0.65, 0.35, and 0.58. We also consider all poisonous clusters as HQ clusters.

After clustering, we now perform CFPM in all clusters separately. We again consider $minsup$ to be 0.5 percent of the dataset size for each class, i.e., $minsup=21.04$ for edible clusters, and $minsup=19.6$ for poisonous clusters. After taking the union over the outputs of CFPM in different clusters, we have 10831 CFPs corresponding to the edible dataset, and 16554 CFPs corresponding to the poisonous dataset. Note that we now have a shorter list of candidate patterns corresponding to each dataset compared to the KRIMP method. 

Using extracted CFPs as candidate patterns, we finally construct two code tables for the edible and the poisonous datasets. The final code table for the edible dataset consists of 183 patterns, and the one for the poisonous dataset consists of 186 patterns. Using constructed code tables via our method, we have
\begin{align*}
\ell(\mathcal{D}_\text{e}\mid CT_\text{e})+\ell(CT_\text{e})=184299\;\; bits,\\
\ell(\mathcal{D}_\text{p}\mid CT_\text{p})+\ell(CT_\text{p})=181261\;\; bits.
\end{align*}
This shows that even after making the list of candidate patterns shorter using our proposed method, we can achieve the same order of compression as the KRIMP method. As the CLOPE algorithm in our method is a low complexity, and fast algorithm, our method makes the process of code table construction computationally much less expensive specially for large transactional datasets.

\subsection{Malware Detection Dataset}
We use the dataset provided by Al-Dujaili et al.~\cite{FGSMMalware} to test our classifier for static malware detection in PE files. Our dataset is constructed using 14772 benign training samples, 14772 malware training samples, 4924 benign test samples, and 4924 malware test samples. The total number of API calls in the dataset is 22761. Therefore, each sample of the dataset is a binary sequence of size 22761 where the locations of ones determine API calls of that sample.

\subsubsection{Neural Network and Its Performance}
We use fully connected feed-forward neural networks to find the state-of-the-art performance for our malware dataset. We use five-fold cross validation to optimize hyper-parameters of our network. Our network consists of five layers: one input layer of size 22761, three hidden layers of size 300, and one output layer of size two. Rectified linear unit (ReLU) is used as the activation function at the hidden layers, and softmax function is used at the output layer. We use drop out rate of 50 percent to avoid over-fitting. The size of mini-batches is 100 samples, the learning rate of Adam optimizer is 0.0001, and the number of epochs is 50. The accuracy, false positive rate (FPR), and false negative rate (FNR) obtained by this network are 91.91, 8.04, and 8.12 percent respectively.

\subsubsection{Proposed Classifier and Its Performance}
We first use the KRIMP method, shown in Fig.~\ref{Fig:TwoStepMethod}, for our dataset. Considering $minsup$ to be 60 percent of the training dataset size for each class, i.e., $minsup=8863.2$, we have 7769 CFPs in the benign training dataset, and 2058 CFPs in the malware training dataset. Using these CFPs as candidate patterns, we construct two code tables for the benign and the malware training datasets referred to as $CT_\text{b}$ and $CT_\text{m}$ respectively. We decide a sample in the test dataset to be malicious if
\begin{equation}\label{eq:testphase}
\ell(\mathcal{T}|CT_\text{m})\leq \ell(\mathcal{T}|CT_\text{b}),
\end{equation}
and to be bengin otherwise. The accuracy, FPR, and FNR obtained by this approach are 85.29, 4.18, and 25.22 percent respectively. In order to improve the performance, we try to use a smaller $minsup$ by which we can extract longer patterns as candidates patterns for model selection. However, by decreasing $minsup$ to be 50 percent of the training dataset size for each class, i.e., $minsup=7386$, we have 218608 CFPs in the benign training dataset, and 85842 CFPs in the malware training dataset. As it can be seen, by decreasing the threshold by only ten percent of the dataset size, we have a dramatic increase in the number of CFPs. This prevents us from using the KRIMP method as we need to set a much smaller $minsup$, and consequently the complexity of both CFPM, and code table construction dramatically increases. 

We therefore use our proposed approach. To cluster the training dataset for each class, we use the CLOPE algorithm with repulsion factor equal to four, and maximum cluster number equal to eight. This provides us with eight clusters for each of the benign, and the malware training datasets. The cluster qualities of the benign training dataset are 0.85, 0.67, 0.25, 0.69, 0.58, 0.84, 0.03, and 0.29. We consider only the cluster with quality 0.03 as a LQ cluster, and consider the rest as HQ clusters. The cluster qualities of the malware training dataset are 0.99, 0.33, 0.89, 0.01, 0.69, 0.36, 0.47, and 0.59. We consider only the cluster with quality 0.01 as a LQ cluster and consider the rest as HQ clusters. 

After clustering, and selecting HQ clusters, we now perform CFPM for HQ clusters separately. We consider $minsup$ to be 0.5 percent of the training dataset size for each class, i.e. $minsup=73.86$. After taking the union, we have 24274 CFPs corresponding to the benign training dataset, and 812 CFPs corresponding to the malware training dataset. Note that we now only have a small number of candidate patterns for each training dataset even by considering a very small $minsup$.

We finally construct two code tables using extracted patterns as candidate patterns for the benign and the malware training datasets. Using these selected models, and the decision criterion in~\eqref{eq:testphase}, the accuracy, FPR, and FNR of our approach are 89.43, 12.77, and 8.36 percent respectively. It can be seen that we have been able to improve the accuracy to be very close to the one for deep neural networks. In the next section, we discuss that MDL-based classifiers can be considered as interpretable classifiers which motivates us to use them even by paying a small penalty in accuracy.

\section{Interpretability of MDL-based Classifiers}\label{Sec:Interpretability}
In this section, we illustrate the interpretability of MDL-based classifiers. As mentioned in the introduction, interpretability is about understanding why a decision is made rather than just what is the decision.  Methods to interpret machine learning models are classified into two classes of intrinsic and post hoc methods. Intrinsic interpretability is when the machine learning model itself is interpretable due to its structure. Post hoc interpretability is when a method is developed to interpret a machine learning model after its training. Machine learning models that are intrinsically interpretable can also be used as a post hoc method by approximating the main model in order to explain its decisions. We here show that MDL-based classifiers can be considered as intrinsically interpretable models. Considering a two-class classifier, we can easily understand why a sample is classified under one class rather than the other one in the following cases.

\textit{Case 1}: The cover of a sample using the code table for one class consists of a few long and several short patterns, and the cover using the code table for the other class consists of many short patterns. This shows that the sample should belong to the class with a few long patterns. This is because longer patterns represent higher similarity with the samples of a dataset. 

\textit{Case 2}: The cover of a sample consists of few patterns considering the code tables for both classes, but the patterns have shorter sum-length under one class, say class~1, than class~2. This says that the cover consists of patterns with higher usage under class~1. Then the sample should be classified under class~1.

\textit{Case 3}: A sample consists of items that their support is zero in one class, but not in the other class. This shows that the sample should not be classified under the class which does not support some of the items.

\section{Conclusion}\label{Sec:Conclusion}
We utilized the minimum description length (MDL) principle, and designed a classifier for large transactional datasets. To do so, we proposed an MDL-based model-selection method for these datasets. The model selection involves first constructing a list of closed frequent patterns (CFPs), and then selecting a subset using the MDL criterion. We showed that, using our method, we can dramatically shorten the list by giving priority to longer patterns as the compression is mainly achieved through longer patterns. This is important as extracting all CFPs, and then summarizing them is computational very expensive due to pattern explosion. We applied our classifier to a dataset of API calls for static malware detection in portable executable (PE) files. We also applied deep neural networks to this dataset to obtain the state-of-the-art performance. The comparison showed that we can obtain an accuracy very close to deep neural networks for our dataset. We also discussed that we can consider MDL-based classifiers as intrinsically interpretable classifiers. Although we might need to pay a small penalty in terms of accuracy, interpretability motivates us to use MDL-based classifiers. We believe that such interpretability can provide a good stepping stone to developing a robust classifier that can withstand evasion attacks. An understanding of why a certain decision is made can provide useful hints about how to counteract functionality-preserving modifications to malware samples.

\section*{Appendix}\label{Sec:Appendix}
In this appendix, we present the two algorithms which have been used in this work for CFPM, and clustering.

\subsection{LCM Algorithm}\label{Sec:LCM}
We here provide an overview of the LCM algorithm~\cite{LCM} for CFPM in transactional datasets. In the LCM algorithm, the closure of a pattern $\mathcal{P}$ is defined~as  
\begin{align*}
Clo(\mathcal{P})=\mathcal{J}(\mathcal{D}(\mathcal{P})),
\end{align*}
where $\mathcal{J}(\mathcal{S})=\cap_{\mathcal{T}_i\in\mathcal{S}} \mathcal{T}_i$. Hence, for every pairs of patterns $\mathcal{P}$ and $\mathcal{Q}$, we have
\begin{itemize}
	
	\item If $\mathcal{P}\subseteq\mathcal{Q}$, then $Clo(\mathcal{P})\subseteq Clo(\mathcal{Q})$,
	\item A pattern $\mathcal{P}$ is closed if and only if $Clo(\mathcal{P})=\mathcal{P}$.
\end{itemize}
In the LCM algorithm, the key notion of prefix preserving closure extension (ppc-extension) is also defined as follows. A pattern $\mathcal{Q}$ is called a ppc-extension of $\mathcal{P}$ if
\begin{itemize}
	\item i) $\mathcal{Q}=Clo(\mathcal{P}\cup\{i\})$ and,
	\item ii) $\mathcal{P}(i-1)=\mathcal{Q}(i-1)$,
\end{itemize}
for some $i\notin\mathcal{P}$ and $i>core_\text{ind}(\mathcal{P})$. $\mathcal{P}(i)=\mathcal{P}\cap\{1,2,\ldots,i\}$, and the core index of $\mathcal{P}$, $core_\text{ind}(\mathcal{P})$, is the minimum index~$j\in\mathcal{P}$ such that $\mathcal{D}(\mathcal{P}(j))=\mathcal{D}(\mathcal{P})$.

Based on the notion of ppc-extension, the LCM algorithm works as follows. It starts with an empty pattern where the core index of an empty set is considered to be zero, i.e., $core_\text{ind}(\{\})=0$. All the frequent ppc-extensions of the empty pattern are calculated. Then, for every newly generated frequent ppc-extension, this procedure is repeated. The algorithm ends when there is no new frequent ppc-extension. All the generated frequent ppc-extensions are considered as the output of this algorithm. It has been proved that this outputs all the CFPs. 

\subsection{CLOPE Algorithm}\label{Sec:CLOPE}

We here provide an overview of the CLOPE algorithm~\cite{clope} as a fast and scalable algorithm for clustering transactional datasets. This algorithm uses the global criterion function defined as
\begin{align}\label{clopeprofit}
Profit=\frac{\sum_{i=1}^{k}\frac{H(\mathbb{C}_i)}{W(\mathbb{C}_i)}|\mathbb{C}_i|}{\sum_{i=1}^{k}|\mathbb{C}_i|}=\frac{\sum_{i=1}^{k}\frac{S(\mathbb{C}_i)}{W^2(\mathbb{C}_i)}|\mathbb{C}_i|}{\sum_{i=1}^{k}|\mathbb{C}_i|},
\end{align}
where $H(\mathbb{C}_i)$, $W(\mathbb{C}_i)$, and $S(\mathbb{C}_i)$ are the height, the width, and the size of cluster~$i$, respectively. The width of a cluster $\mathbb{C}$ is calculated as $W(\mathbb{C})=|\cup_{\mathcal{T}_i\in \mathbb{C}}\mathcal{T}_i|$. The size of a cluster $\mathbb{C}$ is calculated as $S(\mathbb{C})=\sum_{\mathcal{T}_i\in \mathbb{C}}|\mathcal{T}_i|$. Using the width and the size of a cluster $\mathbb{C}$, its height is defined as $H(\mathbb{C})=\frac{S(\mathbb{C})}{W(\mathbb{C})}$.

The global criterion function in~\eqref{clopeprofit} can be generalized as
\begin{align*}
Profit=\frac{\sum_{i=1}^{k}\frac{S(\mathbb{C}_i)}{W^r(\mathbb{C}_i)}|\mathbb{C}_i|}{\sum_{i=1}^{k}|\mathbb{C}_i|},
\end{align*}
where $r$, called repulsion factor, is used to have control over intra-cluster similarity. Larger repulsion factor leads to clusters in which transactions share more common items.

The CLOPE algorithm consists of two phases: allocation phase, and refinement phase. In the allocation phase, we start reading transactions one by one. We either allocate a transaction to an exiting cluster or create a new cluster. This is done to maximise the profit. In the refinement phase, we start reading transactions again. In this phase, we check whether moving a transaction to another existing cluster or a new cluster can increase the profit. If a transaction is moved, we update the clusters, and continue to scan the whole dataset. The algorithm in the refinement phase ends when none of the transactions is moved in an iteration.


\bibliographystyle{IEEEtran}

\begin{thebibliography}{10}
	\providecommand{\url}[1]{#1}
	\csname url@samestyle\endcsname
	\providecommand{\newblock}{\relax}
	\providecommand{\bibinfo}[2]{#2}
	\providecommand{\BIBentrySTDinterwordspacing}{\spaceskip=0pt\relax}
	\providecommand{\BIBentryALTinterwordstretchfactor}{4}
	\providecommand{\BIBentryALTinterwordspacing}{\spaceskip=\fontdimen2\font plus
		\BIBentryALTinterwordstretchfactor\fontdimen3\font minus
		\fontdimen4\font\relax}
	\providecommand{\BIBforeignlanguage}[2]{{%
			\expandafter\ifx\csname l@#1\endcsname\relax
			\typeout{** WARNING: IEEEtran.bst: No hyphenation pattern has been}%
			\typeout{** loaded for the language `#1'. Using the pattern for}%
			\typeout{** the default language instead.}%
			\else
			\language=\csname l@#1\endcsname
			\fi
			#2}}
	\providecommand{\BIBdecl}{\relax}
	\BIBdecl
	
	\bibitem{MLforMalware}
	M.~G. Schultz, E.~Eskin, E.~Zadok, and S.~J. Stolfo, ``Data mining methods for
	detection of new malicious executables,'' in \emph{Proc. IEEE Symp. Security
		and Privacy (S\&P)}, Oakland, USA, May 2001, pp. 38--49.
	
	\bibitem{FGSMMalware}
	A.~{Al-D}ujaili, A.~Huang, E.~Hemberg, and U.~M. O'Reilly, ``Adversarial deep
	learning for robust detection of binary encoded malware,'' in \emph{Proc.
		IEEE Security and Privacy Workshops (SPW)}, San Francisco, USA, May 2018, pp.
	76 -- 82.
	
	\bibitem{StaticNgram}
	J.~Z. Kolter and M.~A. Maloof, ``Learning to detect and classify malicious
	executables in the wild,'' \emph{Journal of Machine Learning Research},
	vol.~7, no.~1, pp. 2721--2744, Dec. 2006.
	
	\bibitem{StaticHeader}
	M.~Z. Shafiq, S.~M. Tabish, F.~Mirza, and M.~Farooq, ``Pe-miner: Mining
	structural information to detect malicious executables in realtime,'' in
	\emph{Proc. Int. Symp. Recent Advances in Intrusion Detection (RAID)},
	Saint-Malo, France, Sep. 2009, pp. 121--141.
	
	\bibitem{EMBER}
	\BIBentryALTinterwordspacing
	H.~S. Anderson and P.~Roth. (2018, Apr. 16) Ember: An open dataset for training
	static pe malware machine learning models. [Online]. Available:
	\url{https://arxiv.org/abs/1804.04637v2}
	\BIBentrySTDinterwordspacing
	
	\bibitem{InterpretableMLBook}
	C.~Molnar, \emph{Interpretable Machine Learning}, 2019,
	\url{https://christophm.github.io/interpretable-ml-book/}.
	
	\bibitem{KRIMP}
	J.~Vreeken, M.~V. Leeuwen, and A.~Siebes, ``Krimp: mining itemsets that
	compress,'' \emph{Data Min. Knowl. Disc.}, vol.~23, no.~1, pp. 169--214, July
	2011.
	
	\bibitem{OddOneOut}
	K.~Smets and J.~Vreeken, ``The odd one out: Identifying and characterising
	anomalies,'' in \emph{Proc. of the 11th SIAM Int. Conf. on Data Mining
		(SDM)}, Mesa, USA, Apr. 2011, p. 804–815.
	
	\bibitem{SLIM}
	------, ``Slim: Directly mining descriptive patterns,'' in \emph{Proc. of the
		12th SIAM Int. Conf. on Data Mining (SDM)}, Anaheim, USA, Apr. 2012, p.
	236–247.
	
	\bibitem{KRIMPCategorical}
	L.~Akoglu, H.~Tong, J.~Vreeken, and C.~Faloutsos, ``Fast and reliable anomaly
	detection in categorical data,'' in \emph{Proc. of the 21st ACM Int. Conf. on
		Information and knowledge management (CIKM)}, Maui, USA, Oct./Nov. 2012, pp.
	415--424.
	
	\bibitem{FPMBook}
	C.~C. Aggarwall and J.~Han, \emph{Frequent Pattern Mining}.\hskip 1em plus
	0.5em minus 0.4em\relax Springer, 2007.
	
	\bibitem{LCM}
	T.~Uno, T.~Asai, Y.~Uchida, and H.~Arimura, ``An efficient algorithm for
	enumerating closed patterns in transaction databases,'' in \emph{Proc. 7th
		international conference on discovery science}, Padova, Italy, Oct. 2004, pp.
	16--31.
	
	\bibitem{Apriori}
	R.~Agrawal and R.~Srikant, ``Fast algorithms for mining association rules,'' in
	\emph{Proc. 20th international conference on very large databases (VLDB)},
	Santiago, Chile, Sep. 1994, pp. 487--499.
	
	\bibitem{LargeItemClustering}
	K.~Wang, C.~Xu, and B.~Liu, ``Clustering transactions using large items,'' in
	\emph{Proc. Eighth Int. Conf. Information and Knowledge Management (CIKM)},
	Kansas City, USA, Nov. 1999, pp. 483--490.
	
	\bibitem{clope}
	Y.~Yang, X.~Guan, and J.~You, ``Clope: A fast and effective clustering
	algorithm for transactional data,'' in \emph{Proc. Eighth ACM SIGKDD Conf.
		Knowledge Discovery and Data Mining (KDD)}, Edmonton, Canada, July 2002, pp.
	682--687.
	
	\bibitem{ITBook}
	T.~A. Cover and J.~A. Thomas, \emph{Elements of Information Theory}.\hskip 1em
	plus 0.5em minus 0.4em\relax John Wiley \& Sons, 2006.
	
\end{thebibliography}
%

\end{document}